\documentclass[sn-vancouver,Numbered]{sn-jnl}% Vancouver Reference Style
\usepackage{graphicx}%
\usepackage{multirow}%
\usepackage{amsmath,amssymb,amsfonts}%
\usepackage{amsthm}%
\usepackage{mathrsfs}%
\usepackage[title]{appendix}%
\usepackage{xcolor}%
\usepackage{soul}
\usepackage{textcomp}%
\usepackage{manyfoot}%
\usepackage{booktabs}%
\usepackage{algorithm}%
\usepackage{algorithmicx}%
\usepackage{algpseudocode}%
\usepackage{listings}%
\setcitestyle{aysep={}}

\usepackage{makecell}
\usepackage{adjustbox}
\newcommand{\ts}{\textsuperscript}
\usepackage[T1]{fontenc}

\begin{document}

\title[Article Title]{Mutual Information Assisted Ensemble Recommender System for Identifying Critical Risk Factors in Healthcare Prognosis}


\author[1]{\fnm{Abhishek} \sur{Dey}}\email{dey.abhishek7@gmail.com}

\author[2]{\fnm{Debayan} \sur{Goswami}}\email{debayang.ju@gmail.com}

\author[3]{\fnm{Rahul} \sur{Roy}}\email{rahul.roy@mahindrauniversity.edu.in}

\author*[2]{\fnm{Susmita} \sur{Ghosh}}\email{susmitaghoshju@gmail.com}

\author[4]{\fnm{Yu Shrike} \sur{Zhang}}\email{yszhang@research.bwh.harvard.edu}

\author[5]{\fnm{Jonathan H.} \sur{Chan}}\email{jonathan@sit.kmutt.ac.th}

\affil[1]{\orgdiv{Department of Computer Science}, \orgname{Bethune College, University of Calcutta}, \state{Kolkata}, \country{India}}

\affil*[2]{\orgdiv{Department of Computer Science and Engineering}, \orgname{Jadavpur University}, \state{Kolkata}, \country{India}}

\affil[3]{\orgname{Mahindra University}, \city{Hyderabad}, \country{India}}

\affil[4]{\orgdiv{Department of Medicine}, \orgname{Brigham and Women's Hospital, Harvard Medical School}, \country{USA}}

\affil[5]{\orgdiv{Innovative Cognitive Computing (IC2) Research Center}, \orgname{King Mongkut’s University of Technology Thonburi}, \country{Thailand}}

\abstract{\textbf{Purpose:} Health recommenders act as important decision support systems, aiding patients and medical professionals in taking actions that lead to patients' well-being. These systems extract the information which may be of particular relevance to the end-user, helping them in making appropriate decisions. The present study proposes a feature recommender, as a part of a disease management system, that identifies and recommends the most important risk factors for an illness. 

\textbf{Methods:} A novel mutual information and ensemble-based feature ranking approach for identifying critical risk factors in healthcare prognosis is proposed.

\textbf{Results:} To establish the effectiveness of the proposed method, experiments have been conducted on four benchmark datasets of diverse diseases (clear cell renal cell carcinoma (ccRCC), chronic kidney disease, Indian liver patient, and cervical cancer risk factors). The performance of the proposed recommender is compared with four state-of-the-art methods using recommender systems' performance metrics like average precision@K, precision@K, recall@K, F1@K, reciprocal rank@K. The method is able to recommend all relevant critical risk factors for ccRCC. It also attains a higher accuracy (96.6\% and 98.6\% using support vector machine and neural network, respectively) for ccRCC staging with a reduced feature set as compared to existing methods. Moreover, the top two features recommended using the proposed method with ccRCC, viz. size of tumor and metastasis status, are medically validated from the existing TNM system. Results are also found to be superior for the other three datasets.

\textbf{Conclusion:} The proposed recommender can identify and recommend risk factors that have the most discriminating power for detecting diseases.}

\keywords{Critical Risk Factors, Feature Recommender, Healthcare Prognosis, Mutual Information, Disease Detection}



\maketitle

\section{Introduction}
In the healthcare domain, the vast amount of clinical data has created an overload of information in terms of medical tests, drugs, and treatment options. This overload leads to indecisiveness in various health-related scenarios both for medical professionals and patients. In such a situation, the application of recommender systems, as a part of a disease management system, lead to accurate and timely patient-related decisions. The healthcare recommender system (HRS) can help the end-users as well as the medical professionals in multiple domains. Example areas include recommendations for diet, drug, and treatment as well as feature recommendations to assist the development of machine learning-based healthcare prognostic systems.

In this article, an automated feature recommender using mutual information with an ensemble strategy is proposed to identify clinically relevant features that can help in disease prognosis. Cervical cancer, chronic kidney disease, liver disease, and clear cell renal cell carcinoma, with high fatality rates, are selected to show the effectiveness of this study. Eight feature selection methods including two wrapper, five filter, and one embedded technique, are used initially to rank the features.  From the ranking information obtained by these eight methods, an ensemble approach is exploited to obtain the final positional feature recommendation table. This positional feature recommendation table contains the features in a ranked manner for further processing. The performance of the proposed method is evaluated using various performance metrics used for recommender systems like average precision@K, precision@K, recall@K, F1@K, and reciprocal rank@K.

The contribution of this work can be highlighted as:
\begin{itemize}
  \item Development of a health recommender system that helps in identifying the critical risk factors for healthcare prognosis. 
  \item An unique ensemble strategy is developed to rank features for the recommendation task. This strategy leverages the strengths of eight different feature selection algorithms, potentially leading to a more robust and informative selection of risk factors compared to using the individual algorithms.
  \item The effectiveness of the proposed method is shown on four standard disease datasets including a detailed demonstration for stage progression of clear cell renal cell carcinoma.
  \item Two supervised classification models are trained using the identified risk factors. The performance of these models is assessed to verify the effectiveness of the recommender system.
\end{itemize}

The rest of the article is organized as follows. Preliminaries for this study and the proposed method is discussed in Section 2. Section 3 includes the datasets, the performance metrics used, the experimental results obtained along with its interpretation. Section 4 concludes the article with closing remarks and the future directions of this research.

\section{Materials and Methods}
This section describes the preliminaries for this study and explains the proposed method.

\subsection{Healthcare Recommender Systems}

With the huge amount of information available online, recommender system acts as a decision support system providing recommendations of items/services which could be useful to the user. This may be in the form of what items to buy, what music to listen, what treatment to undergo \cite{Ricci2010}, etc. Health recommender system is a special area of application in recommender systems where medical items, knowledge, or services are recommended to the end users (patients, medical professionals, etc.) 

\begin{figure}[h!]
    \centering
    \includegraphics[scale=0.4]{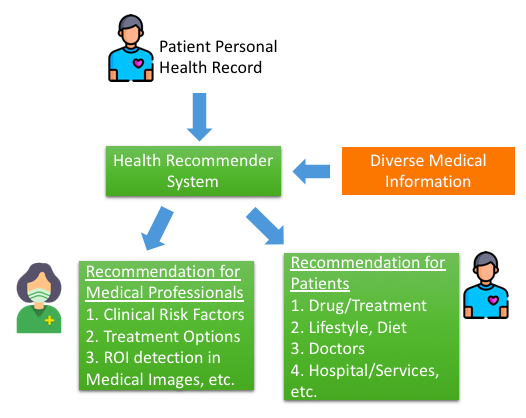}
    \caption{Health Recommender Systems}
    \label{fig:HRS}
\end{figure}

Health recommender systems try to build constructive knowledge out of diverse medical information and combine it with personal health data to make recommendations that can help in healthcare-related decision making as shown in Figure \ref{fig:HRS}. For the patient, this decision support can be in the form of diet/treatment/lifestyle/drug/doctor/hospital recommendation. For the medical professional, this decision support can be in the form of identification of the risk factors of a particular disease, region of interest detection in medical images as well as the recommendation of drug/treatment options.

\subsection{Related Works}
Proper feature selection not only improves the performance but also plays a key factor in bringing explainability to the system. Being able to explain a recommendation or a classification and attributing it to certain features becomes easier when the number of features is less. Thus, feature selection remains an active area of research in the healthcare domain. In \cite{Alhassan2021, rado2019}, the authors performed a comprehensive survey on feature selection for healthcare domain. Multiple studies \cite{ceylan2021, chen2018, halder2009} exist that utilised soft computing-based optimization techniques like genetic algorithm, particle swarm optimization, ant colony optimization for feature selection, and pattern classification. In \cite{Park2016}, wrapper-based feature selection is done using seven common feature selection methods and it is concluded from the experiment that Bayesian network with top 15 features is the most suitable prediction model. However, in \cite{Park2016}, the medical relevance of the selected features is not sufficiently analyzed. Filter and wrapper-based feature selections have been used in many studies \cite{bhavan2018, sahoo2023} for tasks like activity recognition, clinical decision making, classification of medical conditions, etc.

HRS-based decision support system is also becoming an important part of healthcare prognosis. These systems aid better decision-making for both patients as well as healthcare professionals. Both content-based as well as collaborative methods have been explored for the design of HRS \cite{sahoo2019deepreco, waqar2019}. HRS has been proposed for recommending the best doctors \cite{guo2016}, suitable drugs \cite{granda2022}, personalized treatment option \cite{chen2018treatment}, proper diet \cite{ge2015}, managing COVID-19 crisis \cite{sayeb2022} and multiple other areas.

Several works exist in the literature that recommend attributes or deal with feature ranking methods in various domains.  \cite{comp_baczkiewicz2021} presented the use of multi-criteria decision-making for creating a decision support system for e-commerce products. \cite{comp_cataltepe2016} uses various collaborative and content-based recommendation systems for Turkish movie recommendation and also shows how feature selection improves the recommendation results. \cite{comp_kanimozhi2020} builds a recommender system for ranking the most important attributes for breast cancer prognosis using the correlation criterion. \cite{comp_kuanr2023} proposes a feature recommender using multi-criteria decision-making and genetic algorithm for a cervical cancer dataset. \cite{comp_parmezan2021} builds a model for the recommendation of a feature selection algorithm using a meta-learning objective. 

The work presented in this article explores the area of identification and recommendation of clinical risk factors for healthcare prognosis. A feature recommender system has been developed for clear cell renal cell carcinoma staging, then it is generalized for other benchmark datasets. Unlike the existing methods cited above, this article uses an ensemble of multiple feature selection techniques for feature recommendation.

\subsection{Proposed Method}

\begin{figure*}[]
    \centering
    \includegraphics[scale=0.45]{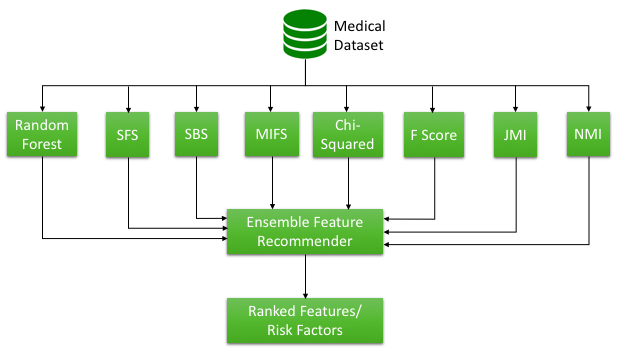}
    \caption{Flow diagram of the proposed method}
    \label{fig:flowchart}
\end{figure*}

Eight feature selection methods have been used for constructing the final positional feature recommendation table, from which the most relevant features may be recommended as shown in Figure \ref{fig:flowchart}. As wrapper methods, Sequential Forward Selection (SFS) and Sequential Backward Selection (SBS) are used because of their implementation simplicity and efficiency in producing a set of features that maximize classification accuracy \cite{datta2011}. Among filter methods, three mutual information(MI)-based (MIFS \cite{Banit2016}, joint mutual information (JMI) \cite{Elmaizi2017} and normalized mutual information (NMI) \cite{Nhaila2019}) have been used in this work. The other two filter methods used are Chi-square and F-score \cite{Remeseiro2019}. As an embedded approach of feature selection, random forest \cite{Gharsalli2016} is used which tends to select a good subset of features that can avoid over-fitting of data without compromising classification accuracy. 

The aim of the proposed article is to identify the features that provide insight into the internal characteristics of data and can also maximize classification accuracy without over-fitting. Such features are expected to be highly ranked by the majority of filter, wrapper, and embedded approaches used in the experiments. Keeping this in mind, a positional table $T$ has been implemented which takes into account the rank of features obtained through the previously mentioned feature selection methods. 

Let the number of features in the dataset be denoted as, $NF$, and the number of feature selection techniques used is denoted as, $NT$. In our experiment, $NT$ is eight. Each location of $T$ can be defined as, $T(i, j)$ = feature index with rank $i$ obtained using feature selection method $j$ ($i$ = 1 to $NF$, $j$ = 1 to $NT$).  Hence, in the table each row corresponds to the rank of the feature and columns indicate feature selection methods. From this table, the importance of any feature is obtained by observing its position. If, for a specific feature all of its positions occur near the top of this table then it can be concluded that the said feature is significant. If, for one particular row ($r$), a single feature ($f$) occurs in the majority of its column positions then it will be selected for inclusion with rank $r$ in the final set. But, if such a single feature is not found then a feature having a higher correlation with the ground truth (i.e., class labels of the dataset) will be selected. This correlation can be measured from the MI-based ranking criterion given below \cite{Novovicova2005}:

\begin{equation}\label{eq2}
MIrank(f) = \sum_{k=1}^{|C|} p(c_k)MI(c_k,f),
\end{equation}

where, $|C|$ is the total number of classes, $p(c_k)$ is the probability of the class $c_k$, $MI(c_k,f)$ is the mutual information between class $c_k$ and feature $f$. Hence, in the absence of a single feature occurring in row $r$ for the majority of times, the feature having the highest MI value with the class labels will be selected with rank $r$. If all the features in row $r$ have already been included in $S$, no new feature can be selected as rank $r$ for that row. In such case, the next selected feature for row $(r+1)$ will be promoted to rank $r$. This will not impose any problem in the feature selection process as it does not violate the ordering of the selection of features. 

\subsubsection{Ensemble-based Feature Recommender}
 
The proposed feature ensemble approach is formally described in Algorithm 1. The process is initialized with $S$ = $\emptyset$ and $R$ = Original set of features in the dataset. $Occ[]$ is an array that is used for storing the number of occurrences of each feature encountered in a single row of positional table $T$ i.e., $Occ[f]$ will give the number of occurrences of feature $f$ in a single row. Evidently, $Occ[f] \leq NT$. $Rank[]$ array stores the final rank of features after the execution of the selection process, $Rank[i]$ will indicate the $i^{th}$ ranked feature ($i$ = 1 to $NF$).

\textbf{Algorithm 1: Pseudocode of the Proposed Algorithm}
\begin{enumerate}
 \item $S$ = $\emptyset$ //Final set of features, initially empty 
 \item $R$ = Original set of features in the dataset
 \item Construct a positional table $T$ so that $T(i, j)$ contains feature index with rank $i$ using feature selection method $j$ ($i$=1 to $NF$, $j$=1 to $NT$)
 \item For $i$=1 to $NF$ do
 \begin{enumerate}
    \item Construct a set $F$ that contains different features with rank $i$ (as obtained from the $i^{th}$ row of $T$)
    \item Calculate the number of occurrences of each feature $f \in F$ and store the value in $Occ[f]$
    \item If (for a specific feature $sf$, ($Occ[sf]$ is higher than other features in $F$) AND ($sf \notin S$))
    \begin{itemize}
        \item $S$=$S$ U ${sf}$ //Updating final set of selected features by including currently selected feature, sf
        \item $R$=$R$ – ${sf}$//Updating remaining set of features by excluding currently selected feature, sf
        \end{itemize}
    \item ElseIf (no such single feature is found) then
    \begin{itemize}
        \item Calculate rank of each feature $f$ such that ($f \in F$) AND ($f \notin S$) based on their MI values using Equation \ref{eq2}
        \item $sf$ = feature with the highest rank obtained from the previous step
        \item $S=S \cup \{sf\}$
        \item $R=R-\{{sf}\}$
    \end{itemize}
    EndIf
    \item $Rank[i]$ = $sf$ //finally $sf$ is selected as the $i^{th}$ ranked feature
 \end{enumerate}
 EndFor
 \end{enumerate}
 
 The above pseudocode provides the sequence of operations to be performed in order to obtain the final rank of features received from the $Rank[]$ array. The 1\ts{st} location of $Rank[]$ contains the feature with the highest rank, the 2\ts{nd} location contains the feature with the 2\ts{nd} highest rank, and so on.  In order to recommend $d$ higher ranked features ($d$ < $NF$), the first $d$ features from the $Rank[]$ array are to be kept in $S$ and other features are to be discarded.

\section{Results and Discussion}
This section explains the datasets and illustrates the results obtained with their interpretation.

\subsection{Datasets}
Experiments have been conducted on four datasets: chronic kidney disease \cite{ckd2015}, Indian liver patient \cite{ilpd2012}, cervical cancer risk factors \cite{cervical2017}, and clear cell renal cell carcinoma. A brief description of these datasets is given in Table \ref{tab:datasets}. Our custom aggregated dataset on ccRCC is obtained from The Cancer Genome Atlas data collections \cite{Akin2016,Clark2013}. The dataset is a combination of clinical, histopathological, demographical, and follow-up for staging of renal cell carcinoma. The dataset consists of twenty-nine features and four classes representing four different stages of renal cancer. A description of these features along with their feature numbers are given in Table \ref{tab:dataset}. For a detailed illustration, the steps of the proposed method are shown on the ccRCC dataset. 

\begin{table}[]
\centering
\caption{Brief description of benchmark datasets}
\label{tab:datasets}
\begin{tabular}{@{}llll@{}}
\toprule
Dataset & \#Features & \#Classes & \#Samples \\ \midrule
Chronic Kidney Disease \cite{ckd2015} & 24 & 2 & 400 \\
Indian Liver Patient \cite{ilpd2012} & 10 & 2 & 583 \\
Cervical Cancer Risk Factors \cite{cervical2017} & 36 & 2 & 858 \\
Clear Cell Renal Cell Carcinoma (ccRCC) & 29 & 4 & 416 \\ \bottomrule
\end{tabular}
\end{table} 

\begin{table*}[]
\centering
\caption{Description of the dataset with feature index number as referred in this article}
\label{tab:dataset}
\resizebox{\textwidth}{!}{\begin{tabular}{@{}llll@{}}
\toprule
\textbf{Feature Number} & \textbf{Name} & \textbf{Description} & \textbf{Type} \\ \midrule
0 & tumor grade & tumor grade G score & Categorical \\
1 & laterality & location of ccRCC kidney & Binary \\
2 & gender & patient’s gender & Binary \\
3 & history other malignancy & history of other tumor disease & Binary \\
4 & history neoadjuvant treatment & history of some chemical therapy & Binary \\
5 & lymph nodes examined count & problem in lymph node & Continuous \\
6 & lymph nodes examined positive & number of cancer nodes & Continuous \\
7 & ajcc tumor pathologic pt & AJCC T phase indicator & Continuous \\
8 & ajcc nodes pathologic pn & AJCC N phase indicator & Categorical \\
9 & ajcc metastasis pathologic pm & AJCC M phase indicator & Binary \\
10 & serum calcium level & calcium level in blood & Categorical \\
11 & haemoglobin level & haemoglobin level in blood & Categorical \\
12 & platelet count & number of platelets in blood 1mm2 & Categorical \\
13 & white cell count & number of white cells in blood 1mm2 & Categorical \\
14 & percent stromal cells & percentage of stromal cells & Continuous \\
15 & percent tumor cell & percentage of tumor cells & Continuous \\
16 & intermediate dimension & dimension related parameter & Continuous \\
17 & longest dimension & dimension related parameter & Continuous \\
18 & shortest dimension & dimension related parameter & Continuous \\
19 & percent necrosis & tumor necrosis percentage & Continuous \\
20 & percent normal cells & percentage normal cells & Continuous \\
21 & percent tumor nuclei & percentage tumor nuclei & Continuous \\
22 & birth days to & lifetime in days & Continuous \\
23 & initial pathologic dx year & initial cancer notification & Continuous \\
24 & lymph nodes examined & number of tumors in lymph node & Categorical \\
25 & vital status & patient’s survival indicator & Categorical \\
26 & last contact days to & number of days to last advise/contact & Continuous \\
27 & tumor status & another kind of tumor except ccRCC & Binary \\
28 & tobacco smoking history indicator & patient's smoking indicator & Categorical \\ \bottomrule
\end{tabular}}
\end{table*}

\subsection{Performance Metrics}
The performance metrics used in this study are precision@K, recall@K, average precision@K, F1@K and reciprocal rank@K \cite{silveira2019}. These metrics assess the quality of recommendation given by the proposed system, where K is the number of recommended items specified by the user. For this article, items refer to the relevant features for prognosis.

\subsection{Positional Feature Recommendation Table}
The positional table obtained for clear cell renal cell carcinoma dataset using Algorithm 1 (as explained in Section 4.1) is shown in Table \ref{tab:featureranks}. From this table, it is evident that feature numbers 7 and 9 are the two most important ones as they are selected by most of the feature selection methods as top-ranked features. Therefore, they are ranked as 1 and 2, respectively, in the final rank. For row numbers: 1, 2, 4, 5, 6, 10, 13, 14, 16, 18, 20, 21, 27, 29 in Table \ref{tab:featureranks}, we observe the occurrence of single features (feature numbers 7, 9, 0, 27, 1, 8, 18, 19, 13, 12, 23, 10, 6, 26, respectively as given in column 3 of Table \ref{tab:mi_featurerank}) in the majority. For other rows, no such single feature is found and the final selection is done using Equation \ref{eq2}. For each feature, MI-based rank is shown in the second column of Table \ref{tab:mi_featurerank}.  Also, for a particular row, already included features in the final set $S$ must not be considered for selection to avoid duplication of features in $S$. It can be observed that for row 23, no feature can be selected because all the features present in row 23 are already included in $S$. Therefore, the next selected feature in row 24 (feature 16) can be considered as rank 23. Note that, it will not violate the ordering of the feature selection process. Once the selection is over, we get 28 features in $S$. Feature number 26 is the one not selected so far. So, it has been included as the last feature (with rank 29) in $S$. In this way a final ranking of features is obtained by the proposed method as shown in the last column of Table \ref{tab:mi_featurerank}.

\begin{table*}[]
\centering
\caption{Ranking of features obtained based on 8 feature selection methods. The numbers in columns 2-9 indicate the feature numbers as shown in Table \ref{tab:dataset}}
\label{tab:featureranks}
\resizebox{\textwidth}{!}{\begin{tabular}{@{}ccccccccc@{}}
\toprule
\textbf{Feature Rank} &
  \multicolumn{8}{c}{\textbf{Feature Selection Method}} \\ \cmidrule(l){2-9} 
 &
  \textbf{Random Forest} &
  \textbf{SFS} &
  \textbf{SBS} &
  \textbf{MIFS \cite{Banit2016}} &
  \textbf{Chi-Squared} &
  \textbf{F-score} &
  \textbf{JMI \cite{Elmaizi2017}} &
  \textbf{NMI \cite{Nhaila2019}} \\ \midrule
1  & 7  & 7  & 7  & 22 & 26 & 9  & 22 & 7  \\
2  & 9  & 9  & 9  & 9  & 7  & 7  & 7  & 9  \\
3  & 27 & 6  & 0  & 7  & 22 & 27 & 26 & 22 \\
4  & 22 & 0  & 15 & 6  & 9  & 0  & 9  & 26 \\
5  & 26 & 15 & 6  & 4  & 5  & 25 & 27 & 27 \\
6  & 0  & 1  & 1  & 20 & 27 & 26 & 0  & 0  \\
7  & 17 & 2  & 2  & 3  & 14 & 13 & 17 & 25 \\
8  & 14 & 3  & 3  & 27 & 19 & 24 & 14 & 5  \\
9  & 16 & 4  & 4  & 28 & 25 & 3  & 25 & 17 \\
10 & 18 & 8  & 8  & 24 & 20 & 17 & 16 & 14 \\
11 & 23 & 10 & 11 & 2  & 24 & 18 & 5  & 19 \\
12 & 5  & 11 & 12 & 12 & 0  & 5  & 15 & 15 \\
13 & 25 & 5  & 18 & 1  & 13 & 11 & 18 & 16 \\
14 & 15 & 17 & 19 & 19 & 6  & 8  & 23 & 13 \\
15 & 21 & 20 & 22 & 11 & 28 & 4  & 19 & 18 \\
16 & 13 & 22 & 10 & 13 & 3  & 28 & 21 & 24 \\
17 & 24 & 13 & 17 & 25 & 4  & 6  & 13 & 20 \\
18 & 12 & 26 & 20 & 10 & 8  & 12 & 24 & 12 \\
19 & 3  & 19 & 21 & 8  & 17 & 1  & 8  & 6  \\
20 & 8  & 21 & 26 & 0  & 23 & 23 & 12 & 3  \\
21 & 10 & 12 & 14 & 21 & 15 & 2  & 10 & 8  \\
22 & 11 & 14 & 24 & 18 & 12 & 15 & 20 & 21 \\
23 & 1  & 18 & 23 & 5  & 1  & 22 & 3  & 23 \\
24 & 28 & 16 & 25 & 23 & 11 & 14 & 6  & 10 \\
25 & 2  & 23 & 27 & 16 & 18 & 19 & 11 & 4  \\
26 & 19 & 24 & 28 & 15 & 2  & 20 & 4  & 11 \\
27 & 4  & 28 & 16 & 17 & 16 & 16 & 28 & 28 \\
28 & 6  & 25 & 13 & 14 & 21 & 21 & 1  & 1  \\
29 & 20 & 27 & 5  & 26 & 10 & 10 & 2  & 2  \\ \bottomrule
\end{tabular}}
\end{table*}

\begin{table}[]
\centering
\caption{Ranking of features obtained based on mutual information (MI) method and our proposed feature selection method. The numbers in columns 2 and 3 indicate the feature numbers as shown in Table \ref{tab:dataset}}
\label{tab:mi_featurerank}
\begin{tabular}{@{}ccc@{}}
\toprule
Feature rank & Mutual Information Based Method & Proposed Method \\ \midrule
1 & 22 & 7 \\
2 & 7 & 9 \\
3 & 26 & 22 \\
4 & 9 & 0 \\
5 & 27 & 27 \\
6 & 0 & 1 \\
7 & 17 & 17 \\
8 & 14 & 14 \\
9 & 25 & 25 \\
10 & 16 & 8 \\
11 & 5 & 5 \\
12 & 15 & 15 \\
13 & 18 & 18 \\
14 & 23 & 19 \\
15 & 19 & 21 \\
16 & 21 & 13 \\
17 & 13 & 24 \\
18 & 24 & 12 \\
19 & 8 & 3 \\
20 & 12 & 23 \\
21 & 10 & 10 \\
22 & 20 & 20 \\
23 & 3 & 16 \\
24 & 6 & 11 \\
25 & 11 & 4 \\
26 & 4 & 28 \\
27 & 28 & 6 \\
28 & 1 & 2 \\
29 & 2 & 26 \\ \bottomrule
\end{tabular}
\end{table}

According to the American Joint Committee on Cancer (AJCC) and Union for International Cancer Control (UICC), there exists a TNM system \cite{Swami2019} to classify the four stages of renal cell carcinoma. In this system, feature numbers: 7(T - tumor size), 8(N - number of lymph nodes examined positive) and 9(M - metastasis status) are the determining factors of the cancer stage. The other relevant risk factors for renal cancer staging as found from literature are feature numbers: 0(tumor grade), 27 (presence of another kind of tumor), 1(location of ccRCC kidney), 17(longest dimension of the tumor) and 5(number of lymph nodes examined) \cite{laber2006,patel2019,ray2016,warren2018}. 

If we consider the top 11 recommendations (K = 11), the proposed method is able to identify all 8 risk factors, depicting better performance with respect to the individual feature selection algorithms used, as shown in Table \ref{tab:comp}. 

\begin{table*}[]
\centering
\caption{Comparison of the proposed method with individual feature selection algorithms used, in terms of the number of relevant features recommended in top 11 recommendations (K = 11) for ccRCC.}
\label{tab:comp}
\resizebox{\textwidth}{!}{\begin{tabular}{@{}lccccccccc@{}}
\toprule
Metric                                      & Proposed Method & Random Forest & SFS   & SBS   & MIFS  & Chi-Squared & F-score & JMI & NMI \\ \midrule
\makecell[l]{Number of \\ relevant features \\ recommended}     & 8               & 5             & 5     & 5     & 3     & 4           & 4       & 6   & 6   \\ \midrule
\makecell[l]{Percentage of \\ relevant features \\ recommended} & 100\%            & 62.5\%         & 62.5\% & 62.5\% & 37.5\% & 50\%         & 50\%     & 75\% & 75\% \\ \bottomrule
\end{tabular}}
\end{table*}

\subsection{Comparison with State-of-the-art Methods on ccRCC dataset}
\begin{figure}[]
    \centering
    \includegraphics[scale=0.36]{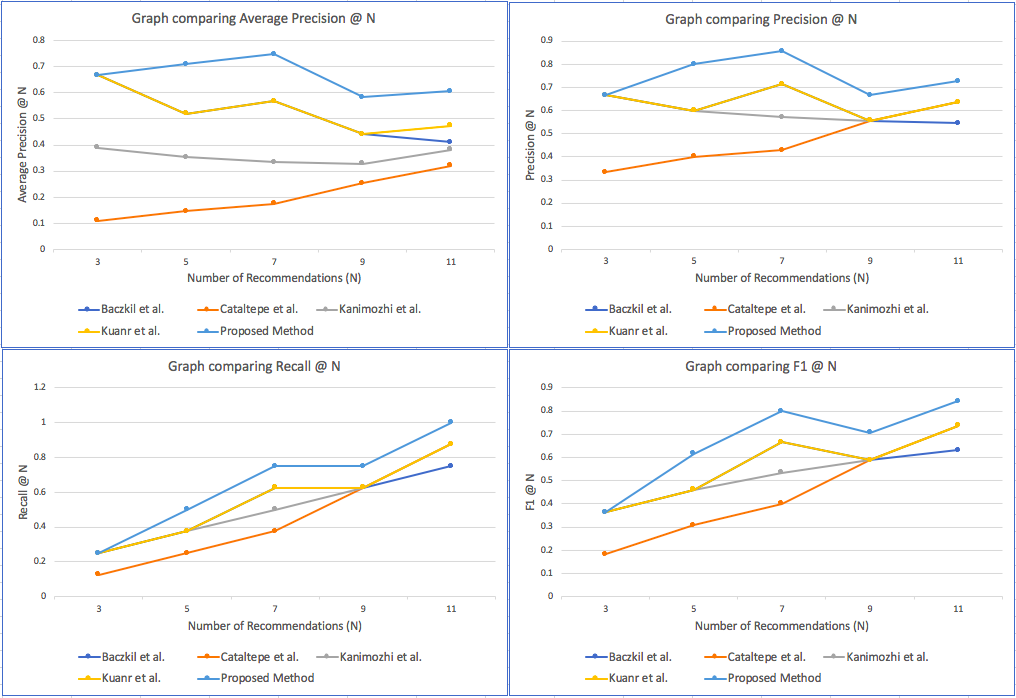}
    \caption{Performance comparison of the proposed method with SOTA methods on ccRCC dataset}
    \label{fig:benchmark}
\end{figure}

The proposed method of feature recommendation has been compared with four state-of-the-art (SOTA) works published between 2016 and 2023 (Baczkiewicz et al. \cite{comp_baczkiewicz2021}, Cataltepe et al. \cite{comp_cataltepe2016}, Kanimozhi et al. \cite{comp_cataltepe2016}, Kuanr et al. \cite{comp_kuanr2023}) on the ccRCC dataset. The results are shown in the Table \ref{tab:comp_benchmark1} and in the form of a graph in Figure \ref{fig:benchmark}. Reciprocal rank (RR) is the same for all K as the rank of the first relevant feature is the same for all K.

The values of the performance metrics are averaged for the values of K = 3,5,7,9,11 in Table \ref{tab:comp_benchmark1}. The individual results for K = 3,5,7,9,11 are shown in the form of a graph in Figure \ref{fig:benchmark}. RR@K is not shown separately in the graph, as it remains the same for all K. It can be seen from Table \ref{tab:comp_benchmark1} and Figure \ref{fig:benchmark} that the proposed method is superior to all the works mentioned above with higher average precision, precision, recall, F1, and equal or better RR.

% It can be seen from Table \ref{tab:comp_benchmark} and Figure \ref{fig:benchmark} that for the number of recommendations (K) = 3, the performance metrics of the proposed method are the same as a few other benchmark methods. However, for K = 5,7,9,11 the proposed method has better average precision@K, precision@K, recall@K, and F1@K than all the benchmark methods used for comparison. The proposed method does a better job of recommending the relevant features when the number of recommendations are high. The performance metrics (average precision@K, precision@K, recall@K averaged for the values of K = 3,5,7,9,11, F1@K and RR@K) are presented in Table \ref{tab:comp_benchmark1} to depict the overall performance. It can be seen that the proposed method, is superior to all the works mentioned above yielding higher average precision, precision, recall, F1, and equal or better RR.

\begin{table*}[htp]
\centering
\caption{Overall comparison of the proposed work with SOTA methods for the ccRCC dataset. Average precision@K, precision@K, and recall@K are averaged for the values of K = 3,5,7,9,11. F1@K is calculated from precision@K and recall@K.}
\label{tab:comp_benchmark1}
\resizebox{\textwidth}{!}{\begin{tabular}{lcllll} 
\toprule
Method                                         & Average Precision@K   & Precision@K & Recall@K & F1@K  & RR@K   \\\midrule
Baczkiewicz et al.\cite{comp_baczkiewicz2021}  & 0.522   & 0.616       & 0.525    & 0.567 & 1.00    \\
Cataltepe et al.\cite{comp_cataltepe2016}      & 0.202   & 0.471       & 0.450    & 0.460 & 0.33 \\
Kanimozhi et al.\cite{comp_kanimozhi2020}      & 0.357   & 0.606       & 0.525    & 0.563 & 0.50  \\
Kuanr et al.\cite{comp_kuanr2023}              & 0.534   & 0.635       & 0.550    & 0.589 & 1.00    \\
\textbf{Proposed method}                       & \textbf{0.663}   & \textbf{0.744}       & \textbf{0.650}    & \textbf{0.694} & \textbf{1.00}    \\
\bottomrule
\end{tabular}}
\end{table*}

\subsection{Performance on Benchmark Datasets}
To show the robustness of the proposed method, the performance of the proposed method is validated on four standard datasets: clear cell renal cell carcinoma(ccRCC), chronic kidney disease \cite{ckd2015}, Indian liver patient \cite{ilpd2012}, and cervical cancer risk factors \cite{cervical2017}. The results obtained are shown in Table \ref{tab:comp_datasets1}.

The proposed method is compared with the next-best method, Kuanr et al. \cite{comp_kuanr2023}, (as in Table \ref{tab:comp_benchmark1}) for the benchmark datasets. The results are shown in Table \ref{tab:comp_2}. The proposed method out-performs Kuanr et al.\cite{comp_kuanr2023} for all the four benchmark datasets.

\begin{table*}[htp]
\centering
\caption{Overall performance of the proposed method on benchmark datasets. Average precision@K, precision@K, and recall@K are averaged for the values of K = 3,5,7,9,11. F1@K is calculated from precision@K and recall@K.}
\label{tab:comp_datasets1}
\resizebox{\textwidth}{!}{\begin{tabular}{lcllll} 
\toprule
Dataset                        & Average Precision@K   & Precision@K & Recall@K & F1@K     & RR@K  \\ \midrule
Cervical Cancer Risk Factors   & 0.892   & 0.913       & 0.443    & 0.597    & 1.00    \\
Chronic Kidney Disease         & 0.669   & 0.755       & 0.318    & 0.448    & 1.00    \\
Clear Cell Renal Cell Carcinoma     & 0.663   & 0.744       & 0.650    & 0.694    & 1.00    \\
Indian Liver Patient Dataset   & 0.864   & 0.872       & 0.725    & 0.792    & 1.00    \\
\bottomrule
\end{tabular}}
\end{table*}

\begin{table*}[htp]
\centering
\caption{Comparison of the proposed method with Kuanr et al., over four datasets. Precision@K and recall@K are averaged for the values of K = 3,5,7,9,11. F1@K is calculated from precision@K and recall@K.}
\label{tab:comp_2}
\resizebox{\textwidth}{!}{\begin{tabular}{lllllllll} 
\toprule
\multirow{2}{*}{Dataset}     & \multicolumn{4}{c}{Proposed Method} & \multicolumn{4}{c}{Kuanr et al. \cite{comp_kuanr2023}}  \\
                             & Precision@K & Recall@K & F1@K    & RR@K    & Precision@K & Recall@K & F1@K    & RR@K  \\ \midrule
Cervical Cancer Risk Factors & 0.913     & 0.443  & 0.597 & 1      & 0.873     & 0.429  & 0.575 & 1    \\
Chronic Kidney Disease       & 0.755     & 0.318  & 0.448 & 1      & 0.646     & 0.271  & 0.381 & 1    \\
Clear Cell Renal Cell Carcinoma   & 0.744     & 0.650  & 0.694 & 1      & 0.635     & 0.550  & 0.589 & 1    \\
Indian Liver Patient Dataset & 0.872     & 0.725  & 0.792 & 1      & 0.872     & 0.400  & 0.548 & 0.5  \\
\bottomrule
\end{tabular}}
\end{table*}

\subsection{Statistical Testing}
The proposed method has been statistically tested against the next-best method in Table \ref{tab:comp_benchmark1}. Table \ref{tab:comp_2} shows the values of the performance metric for the proposed method and Kuanr et al. \cite{comp_kuanr2023} for the four standard datasets. Wilcoxon signed-rank test, as a non-parametric test, has been used for studying the statistical difference between the two methods. This test has been used over the paired T-test as it does not require a normal distribution of the data. More details about the Wilcoxon signed-rank test can be found at \cite{woolson2007}.

The Z-score obtained for the paired observations in Table \ref{tab:comp_2} is -3.02 which is lower than the value of the Z-score for a two-tailed hypothesis with a significance level of 0.05 (-1.96). Thus the difference between the two methods is statistically significant.

\subsection{Classification Results}
As an illustration, the ranking of features obtained from the eight feature selection methods is used to form a recommendation ensemble for the ccRCC dataset. This ensemble creates a positional feature recommendation table shown in column 3 of Table \ref{tab:mi_featurerank}.

Once the final ranking of features is obtained, the two classifiers, NN and SVM, are used to investigate the discriminating power of the selected set of features. A standard NN classifier with 2-hidden layers (14 and 8 nodes) is used to assess how classification accuracy varies with each additional feature. Adam optimizer is used with a learning rate of 0.2. The hyper-parameters are chosen using a grid search methodology. For SVM, the one-vs-all multiclass classification technique is used with the RBF kernel.

The model developed using the features from the proposed feature recommender, for the ccRCC dataset, has also been compared with an existing work by Park et al.\cite{Park2016} that uses clinical features for renal cancer stage classification. In \cite{Park2016}, 15 features were identified to be the most important ones, and they were used to do cancer stage classification. Classification accuracy, for 10-fold crossvalidation, obtained using SVM and NN through \cite{Park2016} and those obtained using our proposed feature selection method are put in Table \ref{tab:comparison_park}. From the table, it is seen that the earlier study yielded an accuracy of 82.3\% and 76.3\% using the SVM and NN classifiers, respectively.  Whereas, our proposed feature recommendation method gives an accuracy of 96.6\% and 98.6\%, respectively with only 4 top-ranked features. This shows the superiority of the model built with features recommended through the proposed approach as compared to that of \cite{Park2016}.

\begin{table}[htp]
\centering
\caption{Comparison with Park et al. \cite{Park2016}}
\label{tab:comparison_park}
\begin{tabular}{@{}ccccc@{}}
\toprule
\textbf{Metric} & \multicolumn{2}{c}{Park et al.\cite{Park2016}} & \multicolumn{2}{c}{Proposed Method} \\ \cmidrule(l){2-5} 
 & SVM & NN & SVM & NN \\ \cmidrule(r){1-5}
Accuracy & 0.823 & 0.763 & \textbf{0.966} & \textbf{0.986} \\ \bottomrule
\end{tabular}
\end{table}

\section{Conclusions}
Proper stage identification or staging is a very important aspect of the treatment process of any disease. Diagnosis of a disease at an early stage can lead to better treatment options. A mutual Information and ensemble based automated feature selection and ranking strategy is proposed in this article. The performance of the proposed method is compared with four state-of-the-art methods and on four different datasets depicting superiority. The proposed method can identify and recommend risk factors that have the most discriminating power for disease diagnosis/staging.  The risk factors recommended for ccRCC by the proposed method can be validated using an existing medical system (AJCC TNM) as well. The results obtained using the proposed method were shown to be statistically significant with respect to the next best method considered for comparison. This work can be extended to study medical images, extracting features, and recommending regions of interest for ease of analysis and diagnosis.

\section{Acknowledgments}
The authors are grateful to the Indo-US Science and Technology Forum (IUSSTF) for establishing a Virtual Networked Center and supporting a part of this research by sanctioning a project titled "Center for Distributed Deep Learning Framework for Classification"   (File No IUSSTF/ JC-024/ 2018 dtd. 13.09.2019). A part of this work has been supported by the IDEAS - Foundation, The TIH at the ISI, Kolkata through sanctioning a Project No. ISI/TIH/2022/55/ dtd. Sept 13, 2022.

\section{Declarations}

\subsection{Competing Interests}
The authors declare that they have no conflicts of interest to report regarding the present study.

\subsection{Availability of Data and Materials}
Data used for this study is available from the corresponding author on request.

\subsection{Author Contributions}
Abhishek Dey, Debayan Goswami, Rahul Roy, and Susmita Ghosh designed the study. Abhishek Dey, Debayan Goswami and Rahul Roy collected the data, conducted the experiments and drafted the manuscript. Results were analyzed by all the authors. Susmita Ghosh, Yu Shrike Zhang, and  Jonathan H. Chan enhanced and finalized the manuscript.

\bibliography{sn-bibliography}% common bib file

\begin{thebibliography}{10}
\providecommand{\doi}[1]{\url{https://doi.org/#1}}
\bibcommenthead

\bibitem[\protect\citeauthoryear{Ricci et~al.}{2010}]{Ricci2010}
Ricci F, Rokach L, Shapira B.
\newblock In: Recommender Systems Handbook. vol. 1-35; 2010. p. 1--35.

\bibitem[\protect\citeauthoryear{Alhassan and Wan~Zainon}{2021}]{Alhassan2021}
Alhassan AM, Wan~Zainon WMN.
\newblock Review of Feature Selection, Dimensionality Reduction and Classification for Chronic Disease Diagnosis.
\newblock IEEE Access. 2021;9:87310--87317.
\newblock \doi{10.1109/ACCESS.2021.3088613}.

\bibitem[\protect\citeauthoryear{Rado et~al.}{2019}]{rado2019}
Rado O, Ali N, Sani HM, Idris A, Neagu D.
\newblock Performance Analysis of Feature Selection Methods for Classification of Healthcare Datasets.
\newblock In: Intelligent Computing. Cham: Springer International Publishing; 2019. p. 929--938.

\bibitem[\protect\citeauthoryear{Ceylan and Atalan}{2021}]{ceylan2021}
Ceylan Z, Atalan A.
\newblock Estimation of healthcare expenditure per capita of {T}urkey using artificial intelligence techniques with genetic algorithm-based feature selection.
\newblock Journal of Forecasting. 2021;40(2):279--290.
\newblock \doi{10.1002/for.2747}.

\bibitem[\protect\citeauthoryear{Chen et~al.}{2018}]{chen2018}
Chen Y, Wang Y, Cao L, Jin Q.
\newblock An Effective Feature Selection Scheme for Healthcare Data Classification Using Binary Particle Swarm Optimization.
\newblock In: 2018 9th International Conference on Information Technology in Medicine and Education (ITME); 2018. p. 703--707.

\bibitem[\protect\citeauthoryear{Halder et~al.}{2009}]{halder2009}
Halder A, Ghosh A, Ghosh S.
\newblock Aggregation pheromone density based pattern classification.
\newblock Fundam Inf. 2009 dec;92(4):345–--362.

\bibitem[\protect\citeauthoryear{Park et~al.}{2016}]{Park2016}
Park K, Ryu K, Ryu K.
\newblock Determining minimum feature number of classification on clear cell renal cell carcinoma clinical dataset.
\newblock In: 2016 International Conference on Machine Learning and Cybernetics (ICMLC). vol.~2; 2016. p. 894--898.

\bibitem[\protect\citeauthoryear{Bhavan and Aggarwal}{2018}]{bhavan2018}
Bhavan A, Aggarwal S.
\newblock Stacked generalization with wrapper-based feature selection for human activity recognition.
\newblock In: 2018 IEEE Symposium Series on Computational Intelligence (SSCI). IEEE; 2018. p. 1064--1068.

\bibitem[\protect\citeauthoryear{Sahoo et~al.}{2023}]{sahoo2023}
Sahoo KK, Ghosh R, Mallik S, Roy A, Singh PK, Zhao Z.
\newblock Wrapper-based deep feature optimization for activity recognition in the wearable sensor networks of healthcare systems.
\newblock Scientific Reports. 2023;13(1):965.

\bibitem[\protect\citeauthoryear{Sahoo et~al.}{2019}]{sahoo2019deepreco}
Sahoo AK, Pradhan C, Barik RK, Dubey H.
\newblock DeepReco: deep learning based health recommender system using collaborative filtering.
\newblock Computation. 2019;7(2):25.

\bibitem[\protect\citeauthoryear{Waqar et~al.}{2019}]{waqar2019}
Waqar M, Majeed N, Dawood H, Daud A, Aljohani NR.
\newblock An adaptive doctor-recommender system.
\newblock Behaviour \& Information Technology. 2019;38(9):959--973.

\bibitem[\protect\citeauthoryear{Guo et~al.}{2016}]{guo2016}
Guo L, Jin B, Yao C, Yang H, Huang D, Wang F, et~al.
\newblock Which doctor to trust: a recommender system for identifying the right doctors.
\newblock Journal of Medical Internet Research. 2016;18(7):e6015.

\bibitem[\protect\citeauthoryear{Granda~Morales et~al.}{2022}]{granda2022}
Granda~Morales LF, Valdiviezo-Diaz P, Re{\'a}tegui R, Barba-Guaman L.
\newblock Drug recommendation system for diabetes using a collaborative filtering and clustering approach: development and performance evaluation.
\newblock Journal of Medical Internet Research. 2022;24(7):e37233.

\bibitem[\protect\citeauthoryear{Chen et~al.}{2018}]{chen2018treatment}
Chen J, Li K, Rong H, Bilal K, Yang N, Li K.
\newblock A disease diagnosis and treatment recommendation system based on big data mining and cloud computing.
\newblock Information Sciences. 2018;435:124--149.

\bibitem[\protect\citeauthoryear{Ge et~al.}{2015}]{ge2015}
Ge M, Ricci F, Massimo D.
\newblock Health-aware food recommender system.
\newblock In: Proceedings of the 9th ACM Conference on Recommender Systems; 2015. p. 333--334.

\bibitem[\protect\citeauthoryear{Sayeb et~al.}{2022}]{sayeb2022}
Sayeb Y, Jebri M, Ghezala HB.
\newblock A graph based recommender system for managing {COVID}-19 Crisis.
\newblock Procedia Computer Science. 2022;196:348--355.

\bibitem[\protect\citeauthoryear{Bączkiewicz et~al.}{2021}]{comp_baczkiewicz2021}
Bączkiewicz A, Kizielewicz B, Shekhovtsov A, Wątróbski J, Sałabun W.
\newblock Methodical Aspects of {MCDM} Based E-Commerce Recommender System.
\newblock Journal of Theoretical and Applied Electronic Commerce Research. 2021;16(6):2192--2229.
\newblock \doi{10.3390/jtaer16060122}.

\bibitem[\protect\citeauthoryear{Çataltepe et~al.}{2016}]{comp_cataltepe2016}
Çataltepe Z, Uluyağmur M, Tayfur E.
\newblock Feature selection for movie recommendation.
\newblock Turkish Journal of Electrical Engineering \& Computer Sciences. 2016;24:833--848.
\newblock \doi{10.3906/elk-1303-189}.

\bibitem[\protect\citeauthoryear{Kanimozhi et~al.}{2020}]{comp_kanimozhi2020}
Kanimozhi G, Shanmugavadivu P, Rani MMS.
\newblock Machine Learning‐Based Recommender System for Breast Cancer Prognosis.
\newblock In: Recommender System with Machine Learning and Artificial Intelligence. 1st ed. India: Wiley; 2020. p. 121--140.

\bibitem[\protect\citeauthoryear{Kuanr and Mohapatra}{2023}]{comp_kuanr2023}
Kuanr M, Mohapatra P.
\newblock Outranking Relations based Multi-criteria Recommender System for Analysis of Health Risk using Multi-objective Feature Selection Approach.
\newblock Data \& Knowledge Engineering. 2023;145:102144.
\newblock \doi{10.1016/j.datak.2023.102144}.

\bibitem[\protect\citeauthoryear{Parmezan et~al.}{2021}]{comp_parmezan2021}
Parmezan ARS, Lee HD, Spolaôr N, Wu FC.
\newblock Automatic recommendation of feature selection algorithms based on dataset characteristics.
\newblock Expert Systems with Applications. 2021;185:115589.
\newblock \doi{10.1016/j.eswa.2021.115589}.

\bibitem[\protect\citeauthoryear{Datta et~al.}{2011}]{datta2011}
Datta A, Ghosh S, Ghosh A.
\newblock Wrapper based feature selection in hyperspectral image data using self-adaptive differential evolution.
\newblock In: 2011 International Conference on Image Information Processing; 2011. p. 1--6.

\bibitem[\protect\citeauthoryear{Banit et~al.}{2016}]{Banit2016}
Banit I, Ouagua NA, Ait~Kerroum M, Hammouch A, Aboutajdine D.
\newblock Band selection by mutual information for hyper-spectral image classification.
\newblock International Journal of Advanced Intelligence Paradigms. 2016 01;8:98.
\newblock \doi{10.1504/IJAIP.2016.074791}.

\bibitem[\protect\citeauthoryear{Elmaizi et~al.}{2017}]{Elmaizi2017}
Elmaizi A, Sarhrouni E, Hammouch A, Nacir C.
\newblock A new band selection approach based on information theory and support vector machine for hyperspectral images reduction and classification.
\newblock In: 2017 International Symposium on Networks, Computers and Communications (ISNCC); 2017. p. 1--6.

\bibitem[\protect\citeauthoryear{Nhaila et~al.}{2019}]{Nhaila2019}
Nhaila H, Elmaizi A, Sarhrouni E, Hammouch A.
\newblock A novel filter approach for band selection and classification of hyperspectral remotely sensed images using normalized mutual information and support vector machines.
\newblock In: Proceedings of EMENA-ISTL 2018; 2019. p. 521--530.

\bibitem[\protect\citeauthoryear{Remeseiro and Bolon-Canedo}{2019}]{Remeseiro2019}
Remeseiro B, Bolon-Canedo V.
\newblock A review of feature selection methods in medical applications.
\newblock Computers in Biology and Medicine. 2019;112:103375.
\newblock \doi{10.1016/j.compbiomed.2019.103375}.

\bibitem[\protect\citeauthoryear{Gharsalli et~al.}{2016}]{Gharsalli2016}
Gharsalli S, Emile B, Laurent H, Desquesnes X.
\newblock Feature Selection for Emotion Recognition based on Random Forest.
\newblock In: Proceedings of the 11th Joint Conference on Computer Vision, Imaging and Computer Graphics Theory and Applications (VISIGRAPP 2016). vol.~4; 2016. p. 610--617.

\bibitem[\protect\citeauthoryear{Novovicova and Malik}{2005}]{Novovicova2005}
Novovicova J, Malik A.
\newblock Information-theoretic feature selection algorithms for text classification.
\newblock In: Proceedings of the International Joint Conference on Neural Networks. vol.~5; 2005. p. 3272--3277.

\bibitem[\protect\citeauthoryear{Rubini et~al.}{2015}]{ckd2015}
Rubini L, Soundarapandian P, Eswaran P.: Chronic kidney disease.

\bibitem[\protect\citeauthoryear{Ramana and Venkateswarlu}{2012}]{ilpd2012}
Ramana B, Venkateswarlu N.: {ILPD} ({I}ndian liver patient dataset).

\bibitem[\protect\citeauthoryear{Fernandes et~al.}{2017}]{cervical2017}
Fernandes K, Cardoso J, Fernandes J.: Cervical cancer (risk factors).

\bibitem[\protect\citeauthoryear{Akin et~al.}{2016}]{Akin2016}
Akin O, Elnajjar P, Heller M, Jarosz R, Erickson BJ, Kirk S, et~al.
\newblock Radiology Data from The Cancer Genome Atlas Kidney Renal Clear Cell Carcinoma [{TCGA-KIRC}] collection.
\newblock The Cancer Imaging Archive. 2016;\doi{10.7937/K9/TCIA.2016.V6PBVTDR}.

\bibitem[\protect\citeauthoryear{Clark et~al.}{2013}]{Clark2013}
Clark K, Vendt B, Smith K, Freymann J, Kirby J, Koppel P, et~al.
\newblock The Cancer Imaging Archive ({TCIA}): Maintaining and Operating a Public Information Repository.
\newblock Journal of Digital Imaging. 2013 12;26/6:1045--1057.

\bibitem[\protect\citeauthoryear{Silveira et~al.}{2019}]{silveira2019}
Silveira T, Zhang M, Lin X, Liu Y, Ma S.
\newblock How good your recommender system is? {A} survey on evaluations in recommendation.
\newblock International Journal of Machine Learning and Cybernetics. 2019;10:813--831.

\bibitem[\protect\citeauthoryear{Swami et~al.}{2019}]{Swami2019}
Swami U, Nussenzveig R, Haaland B, Agarwal N.
\newblock Revisiting {AJCC} {TNM} staging for renal cell carcinoma: {Q}uest for improvement.
\newblock Annals of Translational Medicine. 2019 03;7:S18--S18.
\newblock \doi{10.21037/atm.2019.01.50}.

\bibitem[\protect\citeauthoryear{Laber}{2006}]{laber2006}
Laber DA.
\newblock Risk factors, classification, and staging of renal cell cancer.
\newblock Medical Oncology. 2006;23:443--454.

\bibitem[\protect\citeauthoryear{Patel et~al.}{2019}]{patel2019}
Patel HD, Gupta M, Joice GA, Srivastava A, Alam R, Allaf ME, et~al.
\newblock Clinical stage migration and survival for renal cell carcinoma in the United States.
\newblock European Urology Oncology. 2019;2(4):343--348.

\bibitem[\protect\citeauthoryear{Ray et~al.}{2016}]{ray2016}
Ray RP, Mahapatra RS, Khullar S, Pal DK, Kundu AK.
\newblock Clinical characteristics of renal cell carcinoma: Five years review from a tertiary hospital in Eastern India.
\newblock Indian Journal of Cancer. 2016;53(1):114--117.

\bibitem[\protect\citeauthoryear{Warren and Harrison}{2018}]{warren2018}
Warren AY, Harrison D.
\newblock WHO/ISUP classification, grading and pathological staging of renal cell carcinoma: standards and controversies.
\newblock World Journal of Urology. 2018;36:1913--1926.

\bibitem[\protect\citeauthoryear{Woolson}{2007}]{woolson2007}
Woolson RF.
\newblock Wilcoxon signed-rank test.
\newblock Wiley Encyclopedia of Clinical Trials. 2007;p. 1--3.

\end{thebibliography}

\end{document}